\pdfoutput=1
\documentclass[11pt]{article}

\usepackage[]{emnlp2021}

\usepackage{times}
\usepackage{latexsym}

\usepackage[T1]{fontenc}
\usepackage[utf8]{inputenc}

\usepackage{microtype}

\title{Lingua Custodia's participation at the WMT 2021 Machine Translation \\ using Terminologies shared task}

\author{Melissa Ailem, Jinghsu Liu and Raheel Qader   \\
Lingua Custodia, France\\ 
{\tt \{melissa.ailem,jingshu.liu,raheel.qader\}@linguacustodia.com}  \\}

\begin{document}
\maketitle
\begin{abstract}
This paper describes Lingua Custodia's submission to the WMT21 shared task on machine translation using terminologies. 
We consider three directions, namely English to French, Russian, and Chinese.
We rely on a Transformer-based architecture as a building block, and we explore a method which introduces two main changes to the standard procedure to handle terminologies. The first one consists in augmenting the training data in such a way as to encourage the model to learn a copy behavior when it encounters terminology constraint terms. The second change is constraint token masking, whose purpose is to ease copy behavior learning and to improve model generalization. 
Empirical results show that our method satisfies most terminology constraints while maintaining high translation quality. 
\end{abstract}

\section{Introduction}
Neural-based architectures have become standard for Machine Translation (MT), they are efficient and offer state-of-the-art performance in many scenarios \citep{vaswani2017attention}. However, these models often trained on very large corpora turn out to be less adequate in domains that require very careful use of terminology.     
%Recently, Neural machine Translation (NMT) models \citep{vaswani2017attention} have enjoyed a huge upswing because of their drastic improvement of translation quality compared to Statistical Phrase-Based Translation systems.
%Unfortunately, this  important improvement is paired with losing control about how these translations are generated. In fact, because of their black-box  property,  NMT systems do not provide an explicit source-target link, which makes is difficult to take into account specific terminological constraints when translating domain specific text.
For instance, consider the following English sentence from a biomedical corpus \emph{\textbf{"now for the fever you can take a tachipirina sweet" }}. The term \emph{\textbf{"tachipirina sweet"}}  refers to \emph{\textbf{"paracétamol"}} in French. Unfortunately, a generic English-French Neural MT (NMT) model would translate the above sentence as: \emph{\textbf{"maintenant pour la fièvre tu peux prendre un tachipirina bonbon"}}, where the term \emph{\textbf{"tachipirina sweet"}} is translated \emph{\textbf{"tachipirina bonbon"}}. 

The goal of the WMT21 shared task on machine translation using terminology constraints is to explore methods that can take into account terminology constraints, in order to improve MT models' accuracy and consistency on specific domains. In the literature there are two main families of methods to take into account specific terminologies. One family incorporates terminology constraints at inference \citep{post2018fast, susanto2020lexically}. Members of this category can guarantee strict enforcement of constraints, however this often comes at the cost of higher decoding time and decreased accuracy \citep{hokamp2017lexically, post2018fast}. The other family of method integrates terminologies at training time \citep{dinu2019training, ailem2021encouraging}, and they have the benefit of not changing the NMT model as well as of not incurring additional computational overheads at inference time \citep{crego2016systran, song2019code, dinu2019training}.  

We participate in the following three directions: English to French, Russian, and Chinese, and the system we submit falls into the second family of method incorporating terminologies at training time. More precisely, we explore a variant of the models proposed in \cite{ailem2021encouraging}, which we train for each language pair.
Following this work, we first annotate our training data with the constraints using tags to distinguish constraints terms from other tokens in the sentences. Second, we further perform constraint-token masking, which improves model robustness/generalization as supported by our experiments. 
%We consider 3 language pairs, namely English->French, English->Chinese and English->Russian and we have trained a constraints-grounded NMT model for each language pair. 

The rest of the paper is organized as follows: section \ref{our_approach} reviews the details of our system, section \ref{data} describes the training data selection, the development and test sets, as well as the terminologies used for each language pair, and  section \ref{expe} presents the different experimental settings and results.

{\setlength\tabcolsep{2pt}
\def\arraystretch{1.6}  
\begin{figure*}[h!]
    \centering
    \small
    \begin{tabular}{p{1.5cm}|p{14cm}}
    Source & since COVID-19 shows similarities to \textcolor{blue}{\textbf{SARS-CoV}} and MERS-CoV , it is likely that their effect on pregnancy are similar . \\
    \hline
        Constraints &  \textcolor{blue}{\textbf{SARS-CoV}} $\to$  \textcolor{orange}{\textbf{SARS-CoV}}\\
        \hline
       TADA  &  since COVID-19 shows similarities to $<$S$>$ \textcolor{blue}{\textbf{SARS-CoV}} $<$C$>$ \textcolor{orange}{\textbf{SARS-CoV}} $<$/C$>$ and MERS-CoV , it is likely that their effect on pregnancy are similar . \\
       \hline
        +MASK &  since COVID-19 shows similarities to $<$S$>$ \textcolor{black}{\textbf{MASK}} $<$C$>$ \textcolor{orange}{\textbf{SARS-CoV}} $<$/C$>$ and MERS-CoV , it is likely that their effect on pregnancy are similar . \\
    \end{tabular}
    \caption{Illustration of TrAining Data Augmentation (TADA) and MASK. }
    \label{ex1}
\end{figure*}
}

\section{Method}
\label{our_approach}
Our objective is to encourage neural machine translation to satisfy lexical constraints. To this end, we rely on the approch proposed in \citep{ailem2021encouraging}, which introduces two changes to the standard procedure, namely training data augmentation and token masking. In the following we describe these two operations, which are also depicted in Figures \ref{ex1} and \ref{ex1_1}.
%\subsection{TrAining Data Augmentation (TADA) and token MASKing (MASK)}
%\RQ{be careful, it is a Figure not table}
\paragraph{TrAining Data Augmentation (TADA).} The purpose of this step is to encourage the NMT model to exhibit a copy behavior when it encounters constraint terms whose translation should be consistent with some terminology. This step, illustrated in Figures \ref{ex1} and \ref{ex1_1}, consists in using tags to annotate our training data with the terminology constraints, i.e., indicate the constraints (if any) in a given source sentence. Note that in the literature, there are other variants that use additional information such as source factors \cite{dinu2019training}. We do not use such information, and we specify terminologies using tags only. 
%Similar to previous work, the key idea is to bias the NMT model to exhibit a copy behavior when it encounters constraints.
%To this end, given some source sentence along with some terminology constraints, we use tags to specify the constraints in the source sentence where relevant, as depicted in Figures \ref{ex1} and \ref{ex1_1}. Note that as opposed to previous work, we do not introduce any further information (e.g., source factors), the constraints are specified using tags only.

\paragraph{Token MASKing (MASK).} We further consider masking the source part of the constraint -- tokens in blue -- as illustrated in Figure \ref{ex1} last row. As suggested in \citep{ailem2021encouraging}, this masking strategy provides a more general pattern for the model to learn to perform the copy operation every time it encounters the tag $<S>$ followed by the MASK token. Moreover, this can make the model more apt to support conflicting constraints, i.e., constraints sharing the same source part but which have different target parts. This may be useful in situation in which some tokens must be translated into different targets for some specific documents and contexts at test time. 
%the mask could also help to improve the generalization of the model ... may be to be explored empirically like varying the percentage of constraints in the training data
%\subsection{Mask}

{\setlength\tabcolsep{2pt}
\def\arraystretch{1.6}  
\begin{figure*}[h!]
    \centering
    \small
    \begin{tabular}{p{1.5cm}|p{14cm}}
    Source &  the Canadian government announced CA \$ 275 million in funding for 96 research projects on medical countermeasures against COVID-19 , including numerous  \textcolor{blue}{\textbf{vaccine}} candidates at Canadian universities , with plans to establish a " vaccine bank " of new vaccines for implementation if another Coronavirus outbreak occurs . \\
    \hline
        Constraints &  \textcolor{blue}{\textbf{vaccine}} $\to$  \textcolor{orange}{\textbf{vaccin}},   \textcolor{blue}{\textbf{vaccines}} $\to$  \textcolor{orange}{\textbf{vaccins}},  \textcolor{blue}{\textbf{Coronavirus outbreak}} $\to$  \textcolor{orange}{\textbf{épidémie de coronavirus}} \\
        \hline
       TADA  &  the Canadian government announced CA \$ 275 million in funding for 96 research projects on medical countermeasures against COVID-19 , including numerous  $<$S$>$ \textcolor{blue}{\textbf{vaccine}} $<$C$>$ \textcolor{orange}{\textbf{vaccin}} $<$/C$>$ candidates at Canadian universities , with plans to establish a " $<$S$>$ \textcolor{blue}{\textbf{vaccine}} $<$C$>$ \textcolor{orange}{\textbf{vaccin}} $<$/C$>$ bank " of new $<$S$>$ \textcolor{blue}{\textbf{vaccines}} $<$C$>$ \textcolor{orange}{\textbf{vaccins}} $<$/C$>$ for implementation if another $<$S$>$ \textcolor{blue}{\textbf{Coronavirus outbreak}} $<$C$>$ \textcolor{orange}{\textbf{épidémie de coronavirus}} $<$/C$>$ occurs . \\
       \hline
        +MASK &  the Canadian government announced CA \$ 275 million in funding for 96 research projects on medical countermeasures against COVID-19 , including numerous  $<$S$>$ \textcolor{black}{\textbf{MASK}} $<$C$>$ \textcolor{orange}{\textbf{vaccin}} $<$/C$>$ candidates at Canadian universities , with plans to establish a " $<$S$>$ \textcolor{black}{\textbf{MASK}} $<$C$>$ \textcolor{orange}{\textbf{vaccin}} $<$/C$>$ bank " of new $<$S$>$ \textcolor{black}{\textbf{MASK}} $<$C$>$ \textcolor{orange}{\textbf{vaccins}} $<$/C$>$ for implementation if another $<$S$>$ \textcolor{black}{\textbf{MASK MASK}} $<$C$>$ \textcolor{orange}{\textbf{épidémie de coronavirus}} $<$/C$>$ occurs . \\
    \end{tabular}
    \caption{Illustration of TrAining Data Augmentation (TADA) and MASK (multiple constraints in one sentence). }
    \label{ex1_1}
\end{figure*}
}

\section{Data}
This section provides information and some statistics regarding the datasets for the three language pairs we consider.
\label{data}
%\subsection{Training Data Selection}
\paragraph{Training Data Selection.}
We consider three language pairs, namely English to French, Russian, and Chinese.
Since our method acts at training time, we first perform a training data selection in order to obtain a reasonable number of sentences containing at least one term from the provided terminologies. To do so, we consider both bilingual and monolingual data, provided as part of the shared task. In fact, we observe that bilingual data do not contain many sentences with terminology terms. Thus, we rely on back-translation of monolingual data,  which contains more recent news on COVID-19, to obtain more sentence pairs with terminologies. We rely on OpusMT\footnote{https://github.com/Helsinki-NLP/Opus-MT} to back translate the Russian monolingual to English. For Chinese and French we use in-house translation engines. Note that we further convert the Chinese data into simplified Chinese using OpenCC. Following previous work on terminology control \citep{dinu2019training, ailem2021encouraging}, only 10\% of the training sentences are annotated in order to maintain the model's performance in terminology free cases. 
The details about training data selection for the different language pairs are summarized in tables \ref{enfrdata}, \ref{enrudata} and \ref{enzhdata}.

{\setlength\tabcolsep{1pt}
  \def\arraystretch{1.2}  
\begin{table*}[h!]
    \centering
    \begin{tabular}{|c|c|c|c|}
    \hline
       Data type  & \#sentences  & \#term-grounded sentences & Corpora \\
       \hline
        Monolingual fr & 342,941 & 342,941  & News Crawl 2020\\
        Parallel en-fr &3,110,291& 110,291& NCv16, UN, Common Crawl, Europarl v10\\
       Parallel en-fr (biomedical) &1,733,757& 67,887&  EMEA, Medline Titles, Medline abstracts\\
           \hline
           \cline{1-3}
           \#Total & 5,186,989 & 521,119& \multicolumn{1}{}{} \\
           \cline{1-3}
    \end{tabular}
    \caption{English $\rightarrow$ French data we use for training.}
    \label{enfrdata}
\end{table*}
}

{\setlength\tabcolsep{1pt}
  \def\arraystretch{1.2}  
\begin{table*}[h!]
    \centering
    \begin{tabular}{|c|c|c|p{7cm}|}
    \hline
       Data type  & \#sentences  & \#term-grounded sentences & Corpora \\
       \hline
    Monolingual ru  &997,889 &697,889  & News Commentary, News  \\
       Parallel en-ru  &6,121,064 &3,169 & News Commentary, Wikititles, ParaCrawl, UN, Wikimatrix, Common Crawl, Yandex \\
        Parallel en-ru  (biomedical)  &46,782 & 0  & Medline \\
           \hline
           \cline{1-3}
           \#Total & 7,165,738 & 701,058&  \multicolumn{1}{}{}\\
           \cline{1-3}
    \end{tabular}
    \caption{English $\rightarrow$ Russian data we use for training.}
    \label{enrudata}
\end{table*}
}

{\setlength\tabcolsep{1pt}
  \def\arraystretch{1.2}  
\begin{table*}[h!]
    \centering
    \begin{tabular}{|c|c|c|p{7cm}|}
    \hline
       Data type  & \#sentences  & \#term-grounded sentences & Corpora \\
       \hline
 Monolingual zh   &899,163 & 899,163& News Crawl 2020 \\
    Parallel en-zh (up-sampled) & 12,900  & 12,900 & Wikititles\\
  Parallel en-zh  &6,322,275& 0 & NCv16, ParaCrawl, Wikimatrix, UN, CCMT\\
           \hline
           \cline{1-3}
           \#Total &7,234,338  & 912,063& \multicolumn{1}{}{} \\
           \cline{1-3}
    \end{tabular}
    \caption{English $\rightarrow$ Chinese data we use for training.}
    \label{enzhdata}
\end{table*}
}

%\subsection{Development and Test Sets}
\paragraph{Development and Test Sets.}
For all language pairs, a development and test sets are provided. Note that for the test sets we have access to the source part only. For the dev sets, the terminology constraints associated with each sentence are available, for the test sets this information is not available, and we leverage the terminology files to find constraint terms in these sets. 
%While we annotate the dev sets with the constraint terms associated to each sentence, the terms in the test sets are found automatically using the provided terminology files. 
Just like the training data, both test and  dev sets are augmented with the terminology constraints as presented in figures \ref{ex1} and \ref{ex1_1}.
The dev/test sets of the different language pairs share the same English source file containing 971/2100 sentences respectively. 

%\subsection{Terminologies}
\paragraph{Terminologies.}
For each language pair, we use the provided terminologies to annotate our train, dev and test sets. 
The terminologies consist of respectively 670, 925 and 710 unique source-target terms for English $\rightarrow$ French, Russian and Chinese. We also observe that one source term might be associated with one or more target terms. In that case, when annotating the train and dev sets we choose the target term used in the ground truth translation. For the test set, we select one of the possible terms at random.  

%%%%
{\setlength\tabcolsep{3.7 pt}
  \def\arraystretch{1.6} 
\begin{table*}[h!]
    \centering
    \begin{tabular}{|c|c|c|c|c|c|c|}
    \hline
    {\small \bf Model}  &  {\small \bf BLEU}& {\small \bf Exact-Match Accuracy} & {\small \bf Window Overlap (2)} & {\small \bf Window Overlap (3)} & {\small \bf 1-TERm} & {\small \bf COMET}\\
       \hline
        Transformer &32.12	&0.325&	0.112&	0.114&	0.369	&0.023 \\
          Constrained decoder &40.12&	0.856&	0.306&	0.298&	0.535&	0.416 \\
            TAG+MASK & \textbf{44.90} &	\textbf{0.919}&	\textbf{0.344} &\textbf{0.335} &	\textbf{0.598}&	\textbf{0.681} \\
         \hline
    \end{tabular}
    \caption{Comparison of different models on the English $\rightarrow$ French test set.}
    \label{res}
\end{table*}}	
%%%
{\setlength\tabcolsep{4pt}
\begin{table*}[h!]
    \centering
    \begin{tabular}{|c|c|c|c|c|c|c|}
    \hline
       {\small \bf Language Pair}  &  {\small \bf BLEU}& {\small \bf Exact-Match Accuracy} & {\small \bf Window Overlap (2)} & {\small \bf Window Overlap (3)} & {\small \bf 1-TERm} & {\small \bf COMET}\\
       \hline
        English $\rightarrow$ French & 44.90	&0.919&	0.344	&0.335&	0.598&	0.681 \\
          English $\rightarrow$ Russian &29.13&		0.849&	0.247&	0.248&	0.474&	0.604\\
            English $\rightarrow$ Chinese & 29.16&	0.829&	0.223	&0.225&	0.437&	0.637 \\
         \hline
    \end{tabular}
    \caption{Results of the investigated system (TAG+MASK) across all the language pairs we consider. Results are obtained using the test set.}
    \label{res2}
\end{table*}}
%%%

\section{Experimental results}
\label{expe}
\subsection{Settings}
For English to French and Russian pairs, we first tokenize the terminology files and the train/test/dev sets before annotating them with the terminology constrains. We use the Moses tokenizer \citep{koehn2007moses} for this step.
%we use Moses tokenizer \citep{koehn2007moses} to tokenize the different corpora. 
We then rely on BPE encoding \citep{sennrich2015neural} with 40k merge operations to segment words into subword-units, which results in a joint vocabulary size of 42588 words for English->French, and vocabulary sizes of (44644, 47532) for the (English, Russian) pair. For English->Chinese we rely on sentence piece \citep{kudo2018sentencepiece} for tokenization, which also performs BPE encoding simultaneously and results in a vocabulary size of 52172 for Chinese and 39996 for english. 
%For en->fr  we learn a joint source and target BPE encoding  \citep{sennrich2015neural} with 40k merge operations to segment it into sub-word units, resulting in a vocabulary size of 42588 words. For en->ru we learn a joint BPE encoding  with 40k merge operations, resulting in a vocabulary size of 44644 for english and 47532 for russian. 
%For en-zh we rely on sentence piece \citep{kudo2018sentencepiece} for tokenization and BPE encoding,  we used a joint source and target vocabulary resulting in a vocabulary size of 83172.
We then annotate the train/test/dev sets with the terminology constraints. 

%For en->fr and en->ru, we have first tokenized the training/test/dev sets and terminology files before annotating them with the constraints and then finally apply BPE encoding.
%For en-zh, as tokenization and BPE encoding are done simultaneously with sentence piece, we first apply this latter before annotating the data with the constraints. 

As a building block for our system, we use the transformer architecture  \citep{vaswani2017attention} with 6 stacked encoders/decoders and 8 attention heads. For English-French, the source and target embeddings are tied with the softmax layer. We use 512-dimensional embeddings, 2048-dimensional inner layers for the fully connected feed-forward network and a dropout rate of 0.3.  The models are trained for a minimum of 50 epochs and a maximum of 100 epochs with a batch size of 2000 tokens per iteration and an initial learning rate of $5\times10^{-4}$. For each language pair, the validation set is used to compute the stopping criterion. We use a beam size of 5 during inference for all models.

\subsection{Results}
For all language pairs, the models are evaluated using the standard MT evaluation metrics (BLEU and COMET scores) as well as other terminology-targeted metrics \citep{anastasopoulos2021evaluation}. 
The latter include the "Exact-Match Accuracy" measure, which simply compute the percentage of constraint terms present in the predicted translations. Although this measure provides an indication of terminology satisfaction, it can only assess whether a term is present in the hypotheses without evaluating  whether this target term is correctly placed. To overcome this issue, the authors in \cite{anastasopoulos2021evaluation} proposed an additional measure, namely  "Window Overlap", which computes the percentage of similar tokens surrounding the constraint terms -- within a defined window -- in the ground truth and the generated hypotheses. 
Finally, the models are also evaluated in terms of "Terminology-biased TER" score, which  is an edit distance based metric \citep{snover2006study, anastasopoulos2021evaluation}. 

We compare the our model TAG+MASK with the traditional transformer baseline \citep{vaswani2017attention} and the constrained decoder approach \citep{post2018fast}, which integrates the constraints during inference time. Results on English $\rightarrow$ French data are presented in table \ref{res}. We observe that the TAG+MASK approach significantly improves over baselines in terms of all measures. 

Table \ref{res2} depicts the results that the submitted system reaches across all the language pairs in terms of different metrics.
%The results reached the submitted system across all the language pairs we consider are depicted in 
%Furthermore, table \ref{res2} shows the results obtained for each considered language pair  in terms of all above-mentioned measures. 

\section{Conclusion}
In this paper, we describe our submission to the WMT21 shared task on machine translation using terminologies. We participate in three language pairs, namely English $\rightarrow$ French, Russian and Chinese.
Our system integrates terminology constraints during training by augmenting the data with terminological terms.
Due to the lack of parallel training data containing the terminology terms, we rely on monolingual data for all language pairs to augment the number of sentences containing terminology terms.
Empirical results comparing our approach with terminology grounded as well as terminology free baselines show the effectiveness of the investigated method. 

%In this task, the same terminology is used to annotate  training and test sets. However, to assess the robustness of the model to handle new unseen constraints it would be interesting to test terminology-grounded models on test sets containing constraints never seen during training.

\section*{Acknowledgments}
This work was partially funded by the French Ministry of Defense.

\bibliography{anthology,custom}
\bibliographystyle{acl_natbib}

\appendix

\end{document}